\documentclass[letterpaper, 10 pt, conference]{ieeeconf}  % Comment this line out if you need a4paper

\IEEEoverridecommandlockouts                              % This command is only needed if 
\usepackage{graphicx} % for pdf, bitmapped graphics files
\usepackage{url}
\usepackage{amsmath}
\usepackage{pifont}
\usepackage{stfloats}
\usepackage{tabularx}
\usepackage{caption}
\usepackage{xcolor}

\usepackage{amsbsy}
\usepackage{bm}
\usepackage{algorithm}
\usepackage{algpseudocode}

\usepackage{bbding}

\usepackage{hyperref}
\definecolor{skyblue}{HTML}{009dff}

\hypersetup{
    colorlinks=true,      % Enable colored links
    urlcolor=blue,        % Set color for external hyperlinks (URLs)
}

\newcommand{\eref}[1]{Eq.~\ref{#1}}
\newcommand{\iref}[1]{Fig.~\ref{#1}}
\newcommand{\aref}[1]{Alg.~\ref{#1}}
\newcommand{\tref}[1]{Table.~\ref{#1}}

\title{\LARGE \bf
GeneA-SLAM2: Dynamic SLAM with AutoEncoder-Preprocessed Genetic Keypoints Resampling and Depth Variance-Guided Dynamic Region Removal
}

\author{Shufan Qing$^{1\dag}$,
Anzhen Li$^{1\dag}$,
Qiandi Wang$^{1}$,
Yuefeng Niu$^{1}$,
Mingchen Feng$^{1}$,
Guoliang Hu$^{1}$,
Jinqiao Wu$^{1}$,
\\
Fengtao Nan$^{1}$,
Yingchun Fan$^{1}$\Envelope}

\begin{document}

\maketitle
\thispagestyle{empty}
\pagestyle{empty}

%%%%%%%%%%%%%%%%%%%%%%%%%%%%%%%%%%%%%%%%%%%%%%%%%%%%%%%%%%%%%%%%%%%%%%%%%%%%%%%%
\begin{abstract}
Existing semantic SLAM in dynamic environments mainly identify dynamic regions through object detection or semantic segmentation methods. However, in certain highly dynamic scenarios, the detection boxes or segmentation masks cannot fully cover dynamic regions. Therefore, this paper proposes a robust and efficient GeneA-SLAM2 system that leverages depth variance constraints to handle dynamic scenes. Our method extracts dynamic pixels via depth variance and creates precise depth masks to guide the removal of dynamic objects. Simultaneously, an autoencoder is used to reconstruct keypoints, improving the genetic resampling keypoint algorithm to obtain more uniformly distributed keypoints and enhance the accuracy of pose estimation. Our system was evaluated on multiple highly dynamic sequences. The results demonstrate that GeneA-SLAM2 maintains high accuracy in dynamic scenes compared to current methods.  
Code is available at: \url{https://github.com/qingshufan/GeneA-SLAM2}.
\end{abstract}

%%%%%%%%%%%%%%%%%%%%%%%%%%%%%%%%%%%%%%%%%%%%%%%%%%%%%%%%%%%%%%%%%%%%%%%%%%%%%%%%
\section{Introduction}
\label{sec:introduction}

% --------------------Figure--------------------
\begin{figure*}[ht!]
	\centering
	\includegraphics[width=1.0\textwidth]{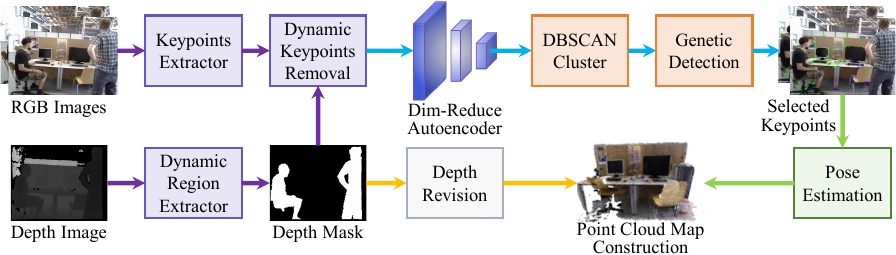}
	\caption{System Overview. GeneA-SLAM2 takes a depth image and two RGB images as input. It first extracts depth-guided dynamic pixels to generate masks and filter out dynamic keypoints (purple arrows). For potentially redundant keypoints, they are dimensionally reduced via an autoencoder, clustered by DBSCAN, and subjected to pose estimation through GeneA sampling (blue arrows). The orange and green arrows show the construction of the global map by stitching point clouds based on estimated poses. }
	\label{fig:overview}
\end{figure*}
% ----------------------------------------------

Simultaneous localization and mapping (SLAM) in dynamic environments plays a critically important role in robot navigation, augmented reality, autonomous driving, and drone trajectory planning. Traditional SLAM systems \cite{ORB3} are primarily designed for static environments, and the presence of a moving object in the environment can significantly affect tracking performance. Common approaches involve introducing motion segmentation and semantic information to eliminate interference from dynamic objects \cite{dyna-slam,ds-slam,cfp-slam}, while some methods rely solely on object detection and depth information to compute semantic information, greatly improving computational efficiency \cite{ngdslam}. However, in some highly dynamic scenarios, object detection cannot always guarantee high accuracy. Even a single frame of failure can lead to artifacts appearing in the point cloud map, seriously interfering with the path planning of unmanned system navigation.  

Depth-based methods have recently gained attention for dynamic object filtering. For example, D2SLAM \cite{d2slam} uses scene depth information to determine the status of keypoints, and DG-SLAM \cite{dgslam} refines semantic information through depth warping, both effectively filtering dynamic points in highly dynamic environments.  However, these methods require semantic prior knowledge and involve complex computations, posing challenges to system real-time performance.  

To address these limitations, we propose GeneA-SLAM2, which extracts dynamic pixels through depth variance and creates precise masks to guide the removal of dynamic objects during tracking and mapping. Additionally, we use autoencoders to reconstruct keypoints, improving the genetic resampling keypoint algorithm in GeneA-SLAM \cite{genea-slam} we proposed earlier to obtain more uniformly distributed keypoints and enhance pose estimation accuracy. Our system was evaluated on multiple highly dynamic sequences, and the constructed global point cloud maps are presented. The results demonstrate that GeneA-SLAM2 maintains high accuracy in dynamic scenes compared to current methods.

Our main contributions are as follows:

\begin{itemize}
	\item A RGB-D SLAM framework, namely GeneA-SLAM2, which can robustly construct point cloud maps with accurate spatial structure and no smear in highly dynamic environments.
	\item A depth mask generation strategy that does not require dynamic object semantic information but only relies on depth variance-aware.
	\item A novel uniform distribution scheme of keypoints based on autoencoder and genetic algorithm to eliminate the phenomenon of feature point clustering.
\end{itemize}

The rest of the paper is organized as follows: Section \ref{sec:related_work} reviews related works. Section \ref{sec:system_description} gives the whole pipeline and details 
of GeneA-SLAM2. Section \ref{sec:experiments} presents the experimental results. Section \ref{sec:conclusion} concludes the paper.  

%%%%%%%%%%%%%%%%%%%%%%%%%%%%%%%%%%%%%%%%%%%%%%%%%%%%%%%%%%%%%%%%%%%%%%%%%%%%%%%%
\section{Related work}
\label{sec:related_work}

\subsection{Traditional Visual SLAM}
A robust and real-time visual SLAM system still relies heavily on feature point computation, with the ORB-SLAM series \cite{ORB3} being a typical example. It is considered the pinnacle of current feature point-based SLAM. Although the quadtree method in ORB - SLAM enhances the uniformity of feature points, there are still some redundant points. These redundant points have an impact on both the accuracy and the speed of pose estimation. PLE-SLAM \cite{ple-slam} introduces line features into point-based visual-inertial SLAM systems, while SiLK-SLAM \cite{silk-slam} improves the learning-based extractor SiLK and introduces a new post-processing algorithm to achieve keypoint homogenization. It is evident that research on the uniformity of feature point extraction is quite important. However, the feature point homogenization strategies of mainstream technologies are still not perfect.

\subsection{Dynamic Visual SLAM}
Since the presence of moving objects often causes ghosting artifacts in constructed maps, significantly degrading mapping accuracy. Eliminating such interference is therefore crucial. Existing methods primarily leverage technologies such as semantic segmentation, object detection, or probabilistic modeling to distinguish dynamic and static objects. D2SLAM \cite{d2slam} models object interactions using depth-related influences, DG-SLAM \cite{dgslam} combines dynamic Gaussian splatting with hybrid pose optimization, Blitz-SLAM \cite{blitz-slam} and DS-SLAM \cite{ds-slam} eliminate dynamic features via semantic information, DynaSLAM \cite{dyna-slam} introduces a dynamic scene inpainting mechanism, and RDS-SLAM \cite{rds-slam} and CFP-SLAM \cite{cfp-slam} achieve real-time dynamic processing based on semantic segmentation and coarse-to-fine probability models, respectively. Additionally, RoDyn-SLAM \cite{rodyn-slam} and ReFusion \cite{refusion} optimize dynamic scene reconstruction using neural radiance fields and residual analysis, while NGD-SLAM \cite{ngdslam} explores GPU-free real-time dynamic SLAM solutions. 

\subsection{Differences from related works}
Different from most dynamic SLAM systems that rely on the accuracy of semantic segmentation to remove dynamic interference, our work focuses on creating depth masks to guide the dynamic regions removal. And the final mask formed by merging the depth mask and NGD-SLAM \cite{ngdslam} mask can completely cover dynamic objects. Our system also inherits the real-time characteristics without GPU from NGD-SLAM.  

In terms of static tracking, GeneA-SLAM2 is similar to our previous work GeneA-SLAM \cite{genea-slam}. To our knowledge, GeneA-SLAM is the first static SLAM system that uses genetic algorithms to optimize the uniform distribution of feature points. However, the feature point homogenization strategy of GeneA-SLAM is not perfect enough to accurately locate redundant points. Unlike GeneA-SLAM \cite{genea-slam}, GeneA-SLAM2 ensures accurate pose estimation and complete dynamic object removal, enabling the construction of point cloud maps that maintain correct spatial geometry and eliminate smears even in highly dynamic environments.

%%%%%%%%%%%%%%%%%%%%%%%%%%%%%%%%%%%%%%%%%%%%%%%%%%%%%%%%%%%%%%%%%%%%%%%%%%%%%%%%
\section{System Description}
\label{sec:system_description}

In this section, we detail the proposed GeneA-SLAM2. Given a depth image of a dynamic environment, dynamic pixels are filtered by statistically calculating the depth value variance through a pixel sliding window, and a mask of dynamic objects is constructed (Sec. \ref{sec:3.1}). To further improve the uniformity of keypoint distribution, this paper clusters the keypoint set reconstructed by an autoencoder and performs genetic algorithm-based resampling of keypoints within each potentially redundant feature point cluster (Sec. \ref{sec:3.2}). Finally, a global point cloud map free from human body regions interference is constructed by revising the depth image through depth constraints (Sec. \ref{sec:3.3}). The system overview of GeneA-SLAM2 is shown in \iref{fig:overview}.

\subsection{Depth-Guided Mask Prediction}
\label{sec:3.1}

Based on the assumption that the variance of human body regions in an indoor environment lies within a specific range, the depth mask of human objects can be calculated. First, a 3$\times$3 sliding window is used to traverse the depth image, recording the depth values of each window while calculating the depth variance:

\begin{equation}
	\label{e1}
	\begin{matrix}
	D_{block}=d(u-1:u+1,v-1:v+1)  \\
	V_{block} = Var(D_{block})
	\end{matrix} 
\end{equation}
where $(u, v)$ represents the image coordinates corresponding to the pixel at the center point of the slider. 

% --------------------Figure--------------------
\begin{figure}[ht!]
	\centering
	\includegraphics[width=0.35\textwidth]{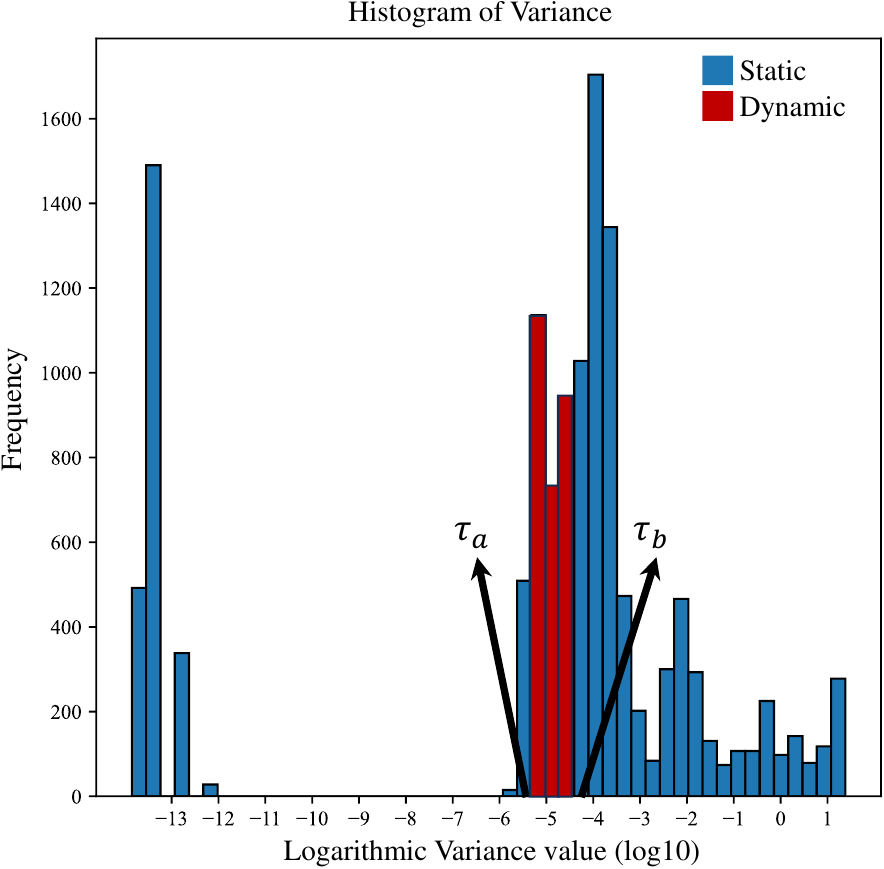}
	\caption{Variance statistical histogram of the window containing dynamic objects. The red bars highlight the variance generated by the potential dynamic parts, i.e., the human body region. By considering the windows whose variances are between $\tau_a$ and $\tau_b$, we can identify the dynamic pixels. }
	\label{fig:hist}
\end{figure}
% ----------------------------------------------

\iref{fig:hist} shows the window variance statistics of a depth image. 
$\tau_a$ and $\tau_b$ are empirical values obtained from experimental statistics. In this paper, $\tau_a = 5e-6$ and $\tau_b = 5e-5$. They are defined to filter windows whose depth variances fall within this range. According to the analysis of Blitz-SLAM for removing depth-unstable regions \cite{blitz-slam}, a depth value of 0 is invalid data. Therefore, for each filtered window, the first pixel with a non-zero depth value within the window is recorded as the feature of the dynamic target. The pseudocode is shown in \aref{alg:dke}.  

\begin{algorithm}[h!]
	\caption{Dynamic pixels extraction algorithm.}
	\label{alg:dke}
	\renewcommand{\algorithmicrequire}{\textbf{Input:}}
	\renewcommand{\algorithmicensure}{\textbf{Output:}}
	\begin{algorithmic}[1]
		\Require
		Current depth image $F_C$.
		\Ensure 
		Pixel set $P_K$.
		\For{$v$ \textbf{from} 1 \textbf{to} $w$ \textbf{step} $3$} //$w$ means width of $F_C$.
		\For{$u$ \textbf{from} 1 \textbf{to} $h$ \textbf{step} $3$} //$h$ means height of $F_C$.
        \State Calc $V_{block}$ according to \eref{e1}
		\If{$\tau_a \leq V_{block} \leq \tau_b$}
		\For{each pixel $(i,j)$ \textbf{in} $D_{block}$}
        \If{$F_C(i,j) > 0$}
		\State $P_K \gets (i,j)$
        \State \textbf{break}
        \EndIf
        \EndFor
		\EndIf
		
		\EndFor
		\EndFor
	\end{algorithmic}
\end{algorithm}

Subsequently, DBSCAN \cite{DBSCAN} is employed to eliminate noise pixels whose depth characteristics are similar to those of dynamic targets, yielding clustering results. 

For each cluster \( C_i \), we first compute the maximum bounding rectangle \( B_i \), then the set of valid depth values within the bounding rectangle can be refined as $\mathcal{D}_i = \{ F_C(x, y) \mid (x, y) \in B_i \text{ and } F_C(x, y) > \tau_c \}$, Next, The local mask \( M_i \) is calculated as:  
\begin{equation}
    M_i(x, y) = 
    \begin{cases}
    1, & |F_C(x, y) - median(\mathcal{D}_i)| \leq \tau_d  \\
    0, & otherwise.
    \end{cases}
\end{equation}
where $\tau_c = 0.05, \tau_d = 0.3$ are empirical values obtained from NGD-SLAM \cite{ngdslam}.

\begin{figure}[htb]
	\centering
	\includegraphics[width=0.4\textwidth]{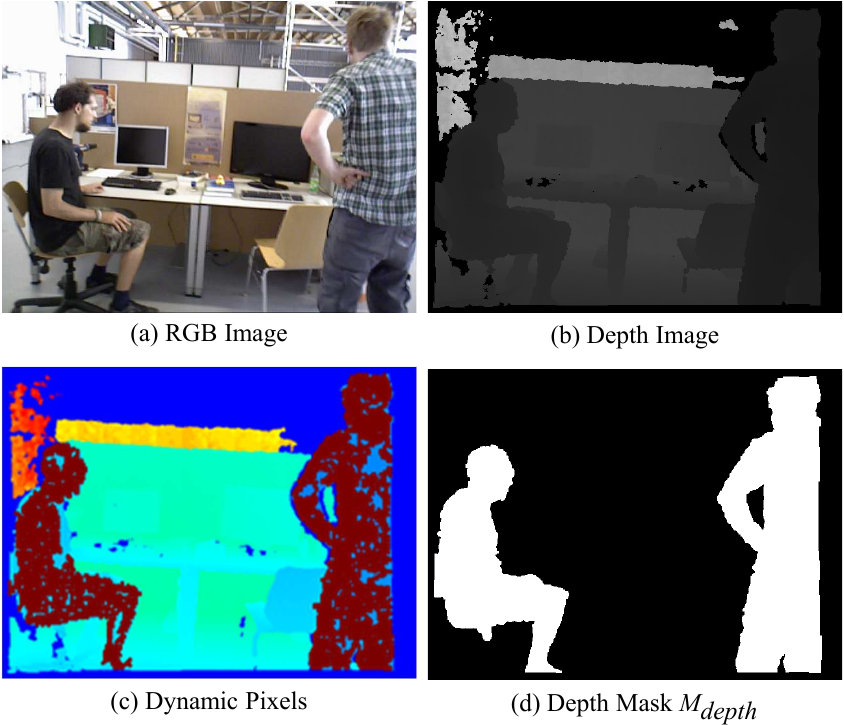}
	\caption{Dynamic Object Depth Mask Prediction.}
	\label{fig:mask-prediction}
\end{figure}

Then, connected component analysis is used to identify the largest connected region in \( M_i \), which is merged into the target mask to obtain the final target mask. Unlike \cite{ngdslam}, this paper provides the most critical dynamic pixels without relying on object detection. An example of obtaining the precise dynamic target mask $M_{depth}$ is shown in \iref{fig:mask-prediction}.

% --------------------Figure--------------------
% ----------------------------------------------

% --------------------Figure--------------------
\begin{figure*}[htb]
	\centering
	\includegraphics[width=0.95\textwidth]{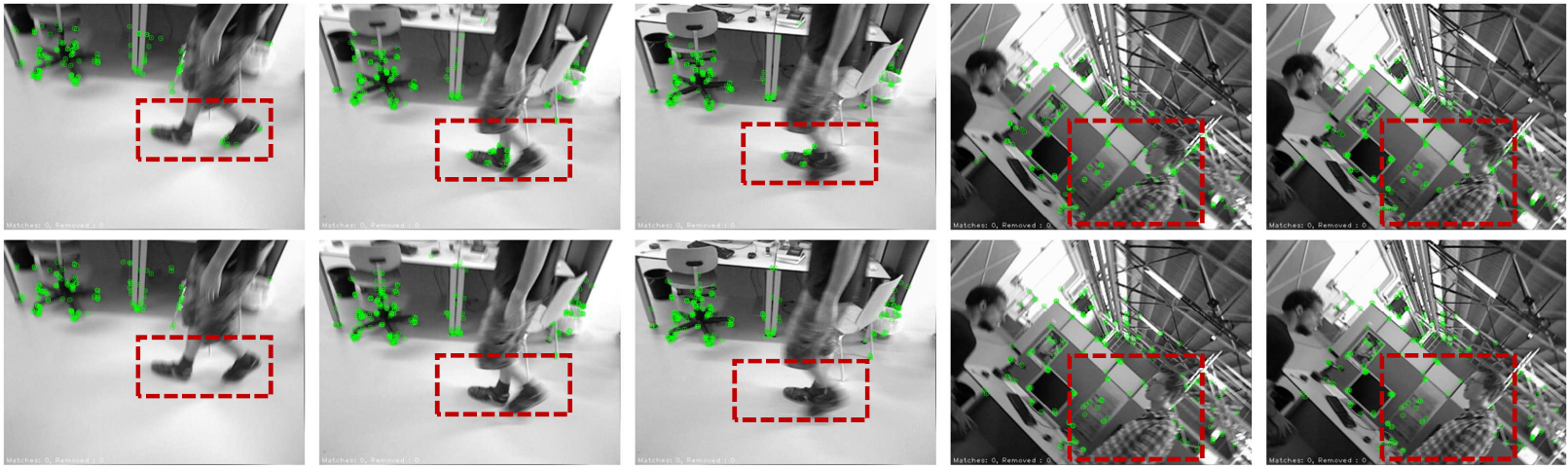}
	\caption{Static Points Tracking. (Top) NGD-SLAM \cite{ngdslam}. (Bottom) GeneA-SLAM2.}
	\label{fig:static-points-tracking}
\end{figure*}
% ----------------------------------------------

When dynamic targets appear in edge regions of images and are incomplete, or their movement speed is too fast, it is extremely likely to cause missed detections by object detection. These missed dynamic feature points will be incorrectly included in the pose estimation calculation process, thereby affecting system accuracy, as shown in  \iref{fig:static-points-tracking}. 

\subsection{Keypoints Resampling based on Genetic Algorithm Optimized by Autoencoder}
\label{sec:3.2}

The main contribution of GeneA-SLAM2 in static tracking is to further enhance the uniformity of keypoint distribution. To achieve this, we use optical flow method to estimate the initial pose. Meanwhile, before solving the PnP problem, we introduce a novel keypoint resampling module guided by an autoencoder \cite{autoencoder}, which only allows the sampled matching points to participate in pose estimation, as shown in \eref{e2}:

\begin{equation}
	\label{e2}
	\mathop{\arg\min}\limits_{\mathbf{R},\mathbf{t}}\sum_{i = 1}^{n} m_i\left\lVert  p_i - \pi\left(\mathbf{R} P_i + \mathbf{t}\right)\right\rVert_2^2
\end{equation}
where $\mathbf{R}$ and $\mathbf{t}$ represent the camera pose to be solved, $(p_i, P_i)$ is a set of 2D-3D matching points, $n$ is the number of matches, and $\pi(\cdot)$ denotes the  projection  that transforms 3D points into 2D pixel coordinates. $m_i = 1$ represents the point is selected, while $m_i = 0$ represents the opposite.

\subsubsection{Keypoint Reconstruction} 
Each keypoint is composed of six characteristic attributes, including 2D coordinates $(x, y)$, diameter $(d)$, main direction $(\theta)$, intensity $(\sigma)$, and pyramid level $(\lambda)$ \cite{opencv_library}. The keypoint set is first reconstructed by an autoencoder network: the encoder maps the original keypoints to a 2D projection space (the latent space dimension defined in this study), and the decoder reconstructs an optimized keypoint set from the projection space. Specifically, a reconstruction loss function based on the L2 norm is used to optimize the autoencoder parameters. This function measures the feature difference by minimizing the Euclidean distance between the original keypoints and the reconstructed keypoints, and its mathematical expression is as follows:

\begin{equation}
	\label{loss}
	\mathcal{L}_k(\boldsymbol{w}, \boldsymbol{u}) = \sum_{i = 1}^n \left\| \boldsymbol{k}_i - g_{\boldsymbol{u}}(f_{\boldsymbol{w}}(\boldsymbol{k}_i)) \right\|_2^2
\end{equation}
where, $\boldsymbol{w}$ is the parameter vector of the encoder, and $\boldsymbol{u}$ is the parameter vector of the decoder. The symbol \(\left\|\cdot\right\|_2^2\) represents the square of the \(L_2\) norm of a vector. The vector $\boldsymbol{k}_i=(x,y,d,\theta,\sigma,\lambda)$ represents the keypoint vector. This vector is first mapped by the encoder $f_{\boldsymbol{w}}$ to obtain the mapped vector $\boldsymbol{k}'_i=f_{\boldsymbol{w}}(\boldsymbol{k}_i)$. Then, it is fed into the decoder $g_{\boldsymbol{u}}$ to reconstruct $\hat{\boldsymbol{k}}_i = g_{\boldsymbol{u}}(\boldsymbol{k}'_i)$. 

\begin{algorithm}[h]
	\caption{Keypoints resampling algorithm.}
	\label{alg:kr}
	\renewcommand{\algorithmicrequire}{\textbf{Input:}}
	\renewcommand{\algorithmicensure}{\textbf{Output:}}
	\begin{algorithmic}[1]
		\Require
		Keypoints set  $K=\{\boldsymbol{k_i}\}_{i=1}^n$, epoch $e = 100$.
		\Ensure 
		Resampling keypoints set  $K_r$.
        \State $q \gets 0.05$ //Initial quantile.
		\For{iter \textbf{from} 1 \textbf{to} $e$}
		\State $K'$ $\gets$ Forward $(K)$ //Encode $K$ to $K'$ and decode $K'$ to $\hat{K}$.
		\State Calc $\mathcal{L}_k(\boldsymbol{w}, \boldsymbol{u})$ according to \eref{loss}
		\State Update $\boldsymbol{w},\boldsymbol{u}$ via gradient descent
		\EndFor
        \State $d_{ij} \gets \left\|\boldsymbol{k}'_i - \boldsymbol{k}'_j \right\|_2^2$
		\State $D = \{ d_{ij} \mid 1 \leq i < j \leq n \}$
        \State Sort $D$ in ascending order.
		\While{$r = 0$ \textbf{and} $ q \leq 0.9$} //Calc $r$
        \State $indicator \gets \lfloor length(D) \cdot q \rfloor$
		\State $r \gets D_{indicator}$ 
		\State $q \gets q + 0.05$
		\EndWhile
		\State $N_{min} \gets \mathrm{dimension}\left(\boldsymbol{k_0}\right)+1$ //Calc $N_{min}$
		\State $C,K_s \gets $ DBSCAN($K'$,$r$,$N_{min}$) //$C$ is the result of clustering, $K_s$ is keypoints that does not require resampling.
		\State $K_{ga} \gets GeneA(C)$ //$K_{ga}$ is the resampling keypoints, refer to our prior work \cite{genea-slam} for details.
		\State $K_r \gets K_s \cup K_{ga}$
	\end{algorithmic}
\end{algorithm}

\subsubsection{Keypoints Resampling} 
\begin{figure}[ht!]
	\centering
	\includegraphics[width=0.45\textwidth]{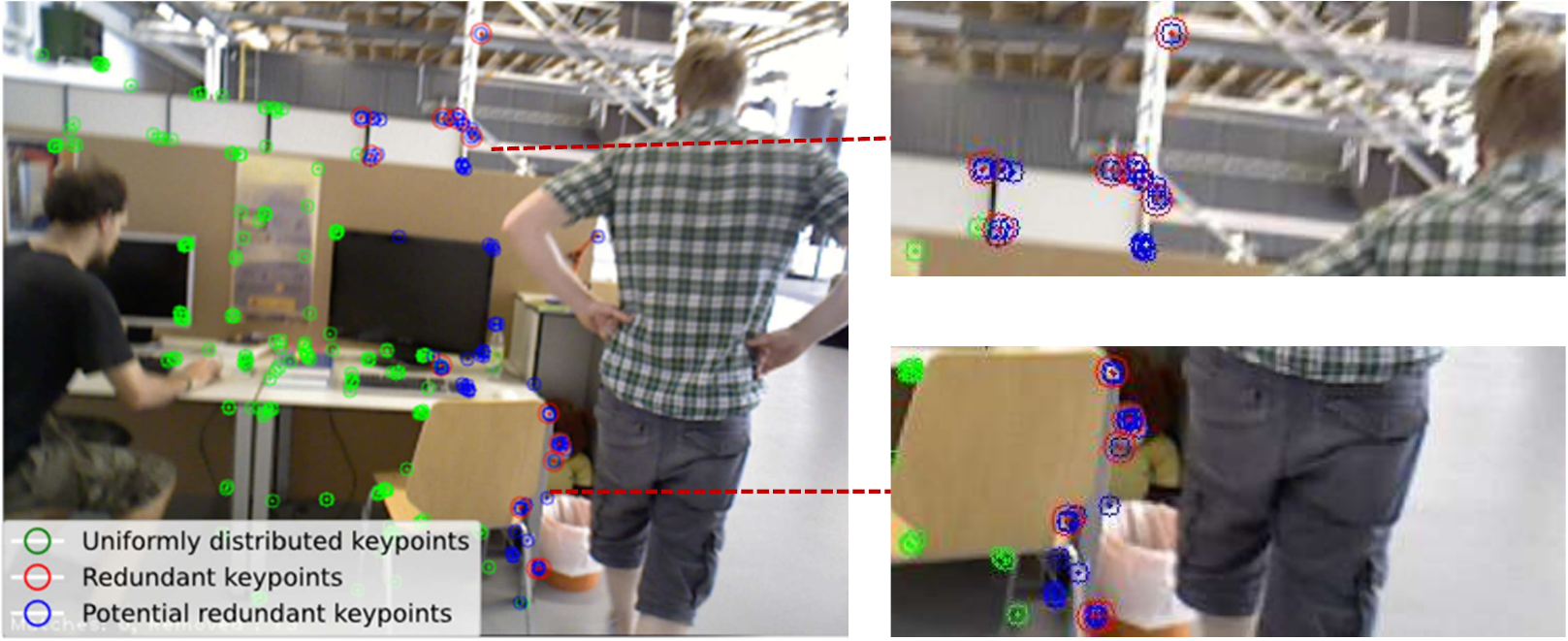}
	\caption{Three types of keypoints. Our method found redundant keypoints (marked with red) in the keypoint clustering regions (marked with blue).}
	\label{fig:keypoints}
\end{figure}

The point set \( K' = \{ \boldsymbol{k}'_i \}_{i = 1}^n \) in the projection space is fed into DBSCAN, which outputs clusters and outlier detection results. Among them, the hyperparameters required for the DBSCAN algorithm, namely the minimum number of points \( N_{\text{min}} \) and the neighborhood radius \( r \), are determined by \aref{alg:kr}. Considering the possible uneven distribution of keypoints in the clusters, we apply the genetic algorithm in GeneA-SLAM \cite{genea-slam} to perform optimization resampling for each cluster. Finally, by integrating the resampling results of all clusters, an optimized keypoint set with uniform distribution is generated, as shown in the blue and green point sets in \iref{fig:keypoints}. Green dots represent uniformly distributed keypoints, blue dots indicate potential redundant keypoints, and red dots are redundant keypoints identified from blue dots. Removing red dots can, to a certain extent, eliminate the aggregation of keypoints in edge and texture regions. And the pseudocode is shown in \aref{alg:kr}.

% --------------------Figure--------------------
% ----------------------------------------------

\subsection{Dynamic Regions Removal of Point Cloud Map
}\label{sec:3.3}

The effectiveness of dense mapping in dynamic environments heavily depends on the accuracy of the mask. In practice, an undersized mask has a far more severe impact than an oversized one. The former may allow information from dynamic objects to leak into the point cloud. To address this, this paper proposes a dynamic object filtering algorithm based on depth constraints. This algorithm deliberately sets a moderately redundant mask to sacrifice some environmental information in exchange for a more precise point cloud map free of dynamic objects.  The specific formula for solving the mask $M_{broad}$ used to filter dynamic objects based on depth constraints is as follows:

\begin{equation}
	M_{broad}(i,j)= 
	\begin{cases}
		0, & \tau_e < F_C(i,j) < \tau_f \\
		1, & otherwise
	\end{cases}
\end{equation}
where $\tau_e = \min_{depth}(P_k)$, $\tau_f = \max_{depth}(P_k)$.
% where, and $P_k$ is the set of dynamic pixels of dynamic objects obtained through \aref{alg:dke}. 

Finally, the mask \( M_{ngd} \) output by the NGD-SLAM mask prediction module is merged with the masks \( M_{depth} \) and \( M_{broad} \) obtained in this paper to derive the final mask \( M_C \):
\begin{equation}
	M_C = M_{ngd} \cup M_{depth} \cup M_{broad}
\end{equation}

% --------------------Figure--------------------
\begin{figure*}[h]
	\centering
	\includegraphics[width=0.95\textwidth]{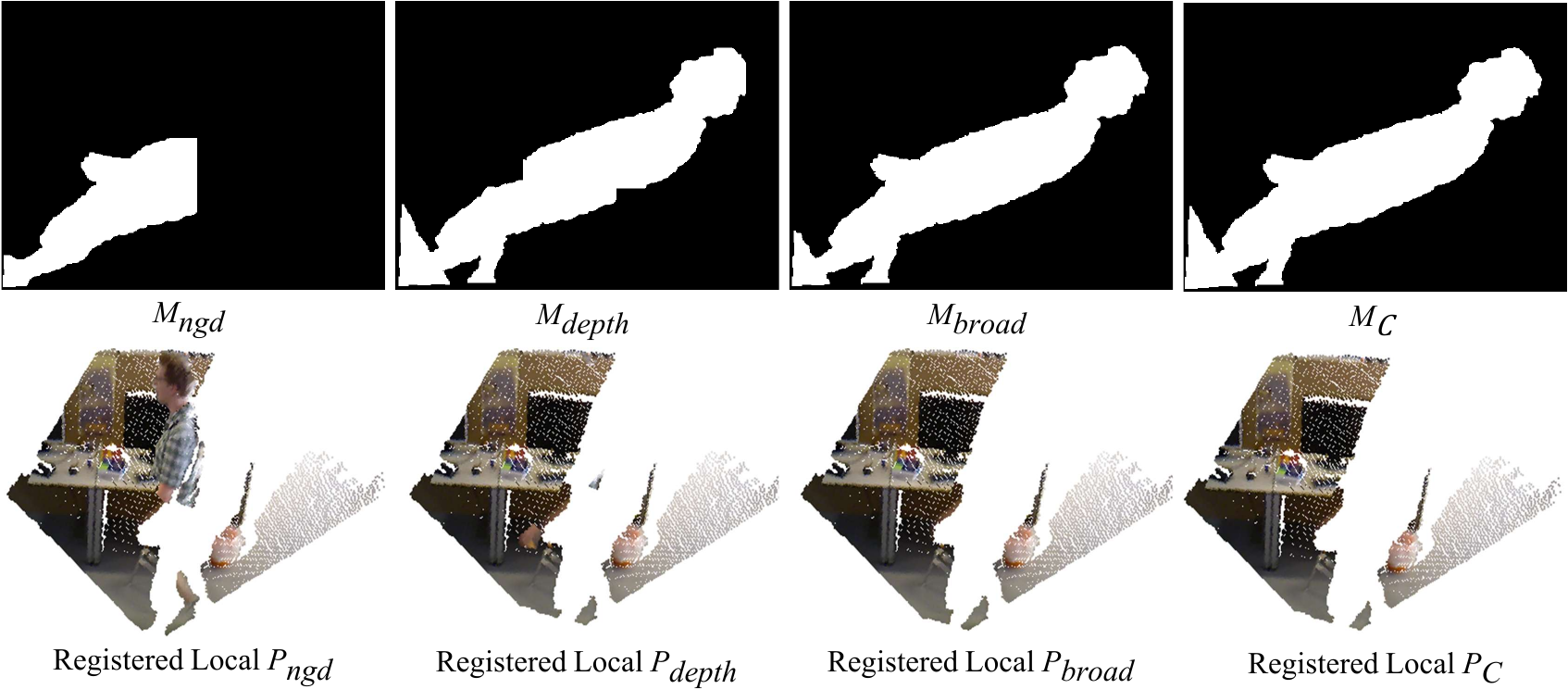}
	\caption{
        Different masks and the corresponding registered local point cloud maps in a frame of the TUM \cite{tum} RGB-D sequence fr3/w/rpy.}
	\label{fig:mask-pcl}
\end{figure*}
% ----------------------------------------------

\iref{fig:mask-pcl} shows that NGD-SLAM cannot effectively handle the rotational motion of the camera in this frame, and $M_{ngd}$ fails to fully cover the human body, resulting in the leakage of partial human body information and contamination of the point cloud map. This is because when the camera rotates at high speed, object detection cannot provide accurate detection boxes. In contrast, our algorithm based on depth constraints can extract more precise masks in such scenarios. The second to fourth columns depict the gradually refined masks and local point cloud maps.

%%%%%%%%%%%%%%%%%%%%%%%%%%%%%%%%%%%%%%%%%%%%%%%%%%%%%%%%%%%%%%%%%%%%%%%%%%%%%%%%
\section{Experiments}
\label{sec:experiments}

\subsection{Experimental Setup}
\subsubsection{Datasets} We utilized all highly dynamic sequences from the TUM RGB-D dataset \cite{tum}, mainly scenarios where two people move around a table. The camera exhibits hemispherical-like trajectories and nearly stationary trajectories, and demonstrates diverse camera motions, including translations and rotations along the XYZ axes (with significant roll, pitch, and yaw changes). Additionally, we employed highly dynamic sequences from the Bonn RGB-D Dynamic Dataset \cite{refusion}, which include two types of sequences: multiple people walking randomly or synchronously, and people carrying static objects. The Bonn dataset shares the same format as the TUM dataset and is captured by a 30Hz depth camera.

\subsubsection{Baseline} In this section, we compare the performance of GeneA-SLAM2 with multiple SLAM methods. In addition to ORB-SLAM3 \cite{ORB3} and NGD-SLAM \cite{ngdslam}, we include DynaSLAM \cite{dyna-slam}, DS-SLAM \cite{ds-slam}, CFP-SLAM \cite{cfp-slam}, RDS-SLAM \cite{rds-slam}, and TeteSLAM \cite{tele-slam}. We also add a dense SLAM method, ACEFusion \cite{acefusion}.

\subsubsection{Metrics} For camera tracking evaluation, we follow the specifications of RGB-D systems and use evo \cite{evo} and SE(3) Umeyama alignment \cite{umeyama} to compute the RMSE of ATE and RPE for the estimated complete camera trajectory \cite{tum}.

\subsubsection{Environment} A laptop computer configured as an AMD Ryzen 5 5600H CPU (6-core 3.30 GHz), 16 GB of RAM, and NVIDIA GeForce RTX 3060 Laptop GPU with 6 GB of memory size.

\subsection{Tracking and Mapping}

% --------------------Figure--------------------
\begin{figure*}[ht!]
	\centering
	\includegraphics[width=0.95\textwidth]{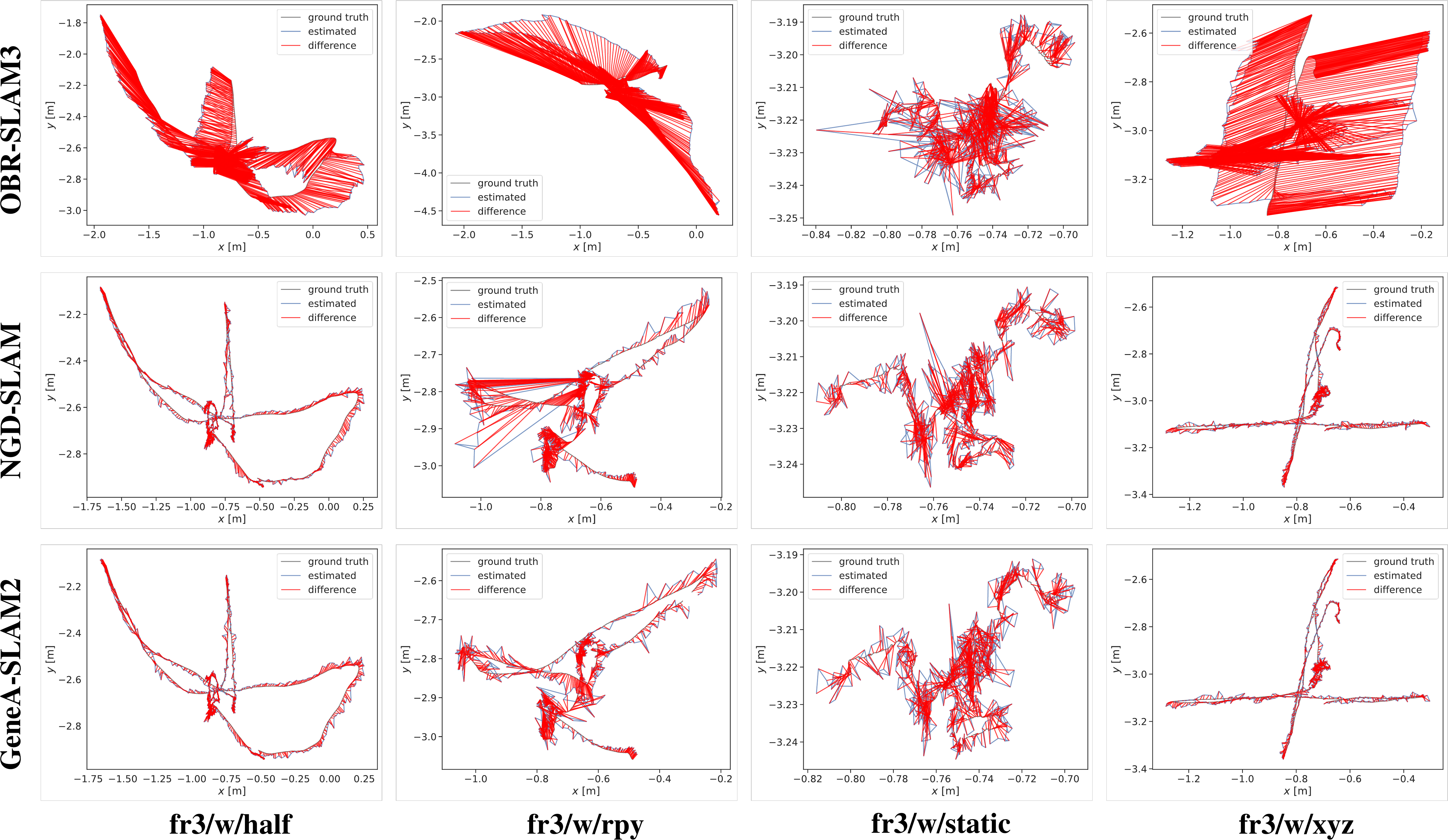}
	\caption{Comparison of the estimated trajectories.}
	\label{fig:tum-traj}
\end{figure*}
% ----------------------------------------------

\begin{table*}[ht!]
	\centering
	\caption{Comparison of RMSE for ATE (m), RPE translation (m/s), and RPE rotation ($^{\circ}$/s) on TUM Highly Dynamic Sequences. The best is boldfaced.}
	\renewcommand{\arraystretch}{1.2}
	\resizebox{\textwidth}{!}{
		\begin{tabular}{|c|ccc|ccc|ccc|ccc|}
			\hline
			Sequences & \multicolumn{3}{c|}{DynaSLAM} & \multicolumn{3}{c|}{DS-SLAM} & \multicolumn{3}{c|}{CFP-SLAM} & \multicolumn{3}{c|}{RDS-SLAM} \\
			\cline{2-13}
			& ATE (m) & RPE (m/s) & RPE ($^{\circ}$/s) & ATE (m) & RPE (m/s) & RPE ($^{\circ}$/s) & ATE (m) & RPE (m/s) & RPE ($^{\circ}$/s) & ATE (m) & RPE (m/s) & RPE ($^{\circ}$/s) \\
			\hline
			f3/w/xyz & 0.015 & 0.021 & 0.452 & 0.025 & 0.033 & 0.826 & \bf{0.014} & 0.019 & 0.602 & 0.057 & 0.042 & 0.922 \\
			f3/w/rpy & 0.036 & 0.045 & 0.902 & 0.444 & 0.150 & 3.004 & 0.037 & 0.050 & 1.108 & 0.160 & 0.132 & 13.169 \\
			f3/w/half & 0.027 & 0.028 & 0.737 & 0.030 & 0.030 & 0.814 & \bf{0.024} & 0.026 & 0.757 & 0.081 & 0.048 & 1.883 \\
			f3/w/static & \bf{0.007} & 0.008 & 0.258 & 0.008 & 0.010 & 0.269 & \bf{0.007} & 0.009 & 0.253 & 0.008 & 0.022 & 0.494 \\
			\hline
			Sequences & \multicolumn{3}{c|}{TeteSLAM} & \multicolumn{3}{c|}{NGD-SLAM} & \multicolumn{3}{c|}{GeneA-SLAM2(\textbf{Ours})} & \multicolumn{3}{c|}{Imp.} \\
			\cline{2-13}
			& ATE (m) & RPE (m/s) & RPE ($^{\circ}$/s) & ATE (m) & RPE (m/s) & RPE ($^{\circ}$/s) & ATE (m) & RPE (m/s) & RPE ($^{\circ}$/s) & ATE (m) & RPE (m/s) & RPE ($^{\circ}$/s) \\
			\hline
			f3/w/xyz & 0.019 & 0.023 & 0.636    & 0.015 & 0.020 & 0.470 & 	 	\bf{0.014} & \bf{0.011} & \bf{0.381} & 		\bf{6.67\%	}&	\bf{45.00\%}		&	\bf{18.94\%} \\
			f3/w/rpy & 0.037 & 0.047 & 1.058    & 0.034 & 0.044 & 0.889 & 	 	\bf{0.030} & \bf{0.021} & \bf{0.493} & 		\bf{11.76\%	}&	\bf{52.27\%}		&	\bf{44.54\%} \\
			f3/w/half & 0.029 & 0.042 & 0.965   & \bf{0.024} & 0.025 & 0.695 & 0.025 & \bf{0.013} & \bf{0.402} 	  & 	-4.17\%	&	\bf{48.00\%}		&	\bf{42.16\%} \\
			f3/w/static & 0.011 & 0.011 & 0.287 & \bf{0.007} & 0.009 & 0.262 & \bf{0.007} & \bf{0.006} & \bf{0.168}  & 	0.00\%	&	\bf{33.33\%}		&	\bf{35.88\%} \\
			\hline
		\end{tabular}
	}
	\label{tab:accuracy-cmp-tum}
\end{table*}

\begin{table}[ht!]
	\begin{center}
		\caption{Comparison of RMSE for ATE (m) on BONN Dataset. The best is boldfaced.}
		\label{tab:accuracy-cmp-bonn}
		\renewcommand{\arraystretch}{1.0}
		%\resizebox{\linewidth}{!}{
			\scriptsize
			\begin{tabular}{|c|c|c|c|c|c|c|c|}
				\hline
				Sequences 		& Dyna  & ACEFusion   & NGD 	& Ours & Imp. \\
				\hline
				crowd 			& \bf{0.016}& \bf{0.016}  & 0.024 		& 0.017 		& \bf{29.17\%  }\\
				crowd2 			& 0.031 	& 0.027 	  & 0.025 		& \bf{0.020} 	& \bf{20.00\%   }\\
				crowd3 			& 0.038 	& \bf{0.023}  & 0.033 		& 0.028 		& \bf{15.15\%  }\\
				mov/no/box 		& 0.232 	& 0.070 	  & \bf{0.016}	& 0.021			& -31.25\%\\
				mov/no/box2 	& 0.039 	& \bf{0.029}  & 0.036 		& 0.030 		& \bf{16.67\%  }\\
				person/track 	& 0.061 	& 0.070 	  & 0.046 		& \bf{0.044} 	& \bf{4.35\%   }\\
				person/track2 	& 0.078 	& 0.071 	  & 0.062 		& \bf{0.055} 	& \bf{11.29\%  }\\
				synchronous 	& 0.015 	& 0.014 	  & 0.028 		& \bf{0.012} 	& \bf{57.14\%  }\\
				synchronous2 	& 0.009 	& 0.010 	  & 0.009 		& \bf{0.008} 	& \bf{11.11\%  }\\
				\hline
			\end{tabular}%
			%}
	\end{center}
\end{table}

\iref{fig:tum-traj} shows the estimated trajectories of three SLAM systems in four highly dynamic sequences. The black line represents the ground truth, the blue line shows the estimated trajectory, and the red line indicates the difference between the ground truth and the estimated trajectory. It can be visually observed from the first row that ORB-SLAM3 cannot effectively handle highly dynamic environments. Compared with ORB-SLAM3, the camera trajectory estimation errors of the other two SLAM systems are significantly reduced.

The comparison results between GeneA-SLAM2 and various SLAM baseline systems on the TUM highly dynamic dataset \cite{tum} are shown in \tref{tab:accuracy-cmp-tum}. In particular, the improvement ratio of GeneA-SLAM2 relative to NGD-SLAM is presented in the last column of the table. Notably, the GeneA-SLAM2 system demonstrates high precision in the four dynamic sequences, followed by CFP-SLAM and NGD-SLAM. However, the CFP-SLAM and NGD-SLAM all perform poorly in the fr3/w/rpy sequence, as can also be seen from the trajectory error of NGD-SLAM in \iref{fig:tum-traj}. This may be because poor object detection results during high-speed camera rotation lead to some dynamic points participating in pose estimation, causing errors.  

The evaluation results on the BONN RGB-D Dynamic Dataset are shown in \tref{tab:accuracy-cmp-bonn}. We used partial experimental data from \cite{ngdslam}, and the results show that our system achieves the highest accuracy in five sequences, followed by ACEFusion, NGD-SLAM, and DynaSLAM, demonstrating the system's applicability in a broader range of dynamic environments. Experiments on BONN show that our system has a significant improvement over NGD-SLAM, with the highest RMSE of ATE improvement of 57.14\% achieved on the synchronous sequence of BONN.

% --------------------Figure--------------------
\begin{figure*}[ht!]
	\centering
	\includegraphics[width=0.95\textwidth]{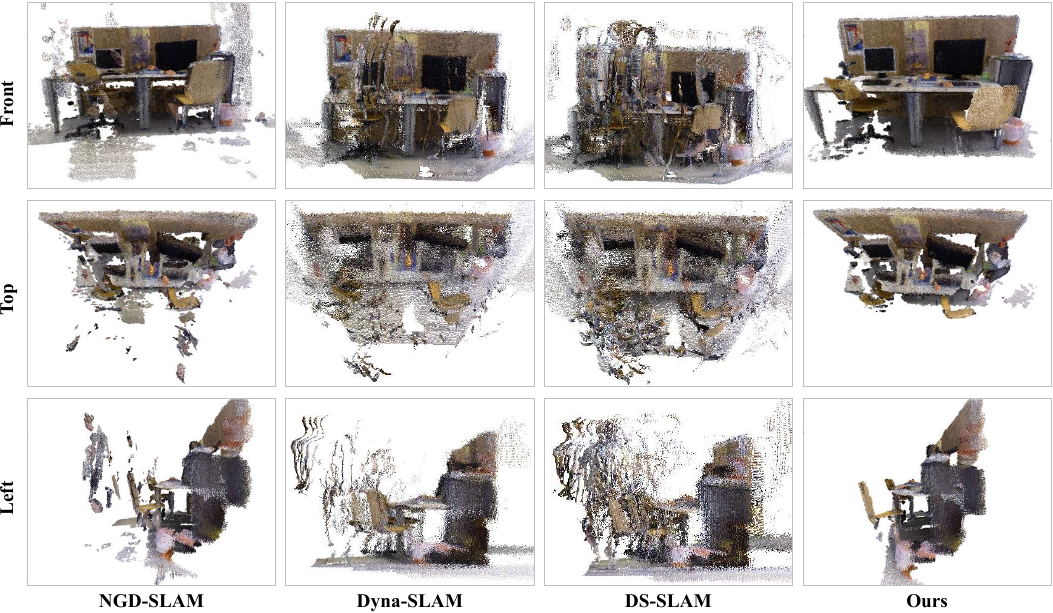}
	\caption{Comparison of the point cloud maps constructed by the four systems in the highly dynamic sequences fr3/w/xyz. Our GeneA-SLAM2 can give the clearest point cloud map.}
	\label{fig:map-walking}
\end{figure*}
% ----------------------------------------------

% --------------------Figure--------------------
\begin{figure*}[ht!]
	\centering
	\includegraphics[width=0.95\textwidth]{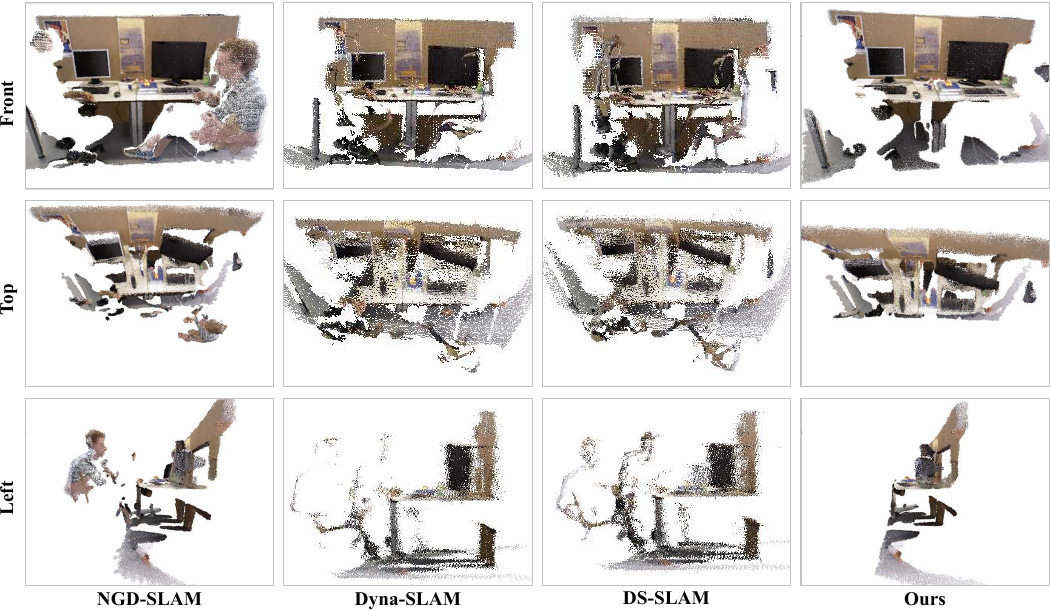}
	\caption{Comparison of the point cloud maps constructed by the four systems in the lowly dynamic sequences fr3/s/static. Our GeneA-SLAM2 can give the clearest point cloud map.}
	\label{fig:map-static}
\end{figure*}
% ----------------------------------------------

\iref{fig:map-walking} displays three views of the global point cloud maps obtained by four SLAM systems. NGD-SLAM, Dyna-SLAM, and DS-SLAM remove the interference from two people. Therefore, the overall spatial geometry of the map is well maintained. But due to the imperfection of dynamic object masks, partial information of the two people leaks into the point cloud maps. Although the noise blocks in the point cloud maps constructed by the three systems are relatively sparse and the objects in the environment can be clearly seen, it is difficult for unmanned system navigation to determine whether the space occupied by these noise blocks is passable. The point cloud map obtained by GeneA-SLAM2 has almost no residual information of people. 

The challenge in mapping low-dynamic sequences lies in the difficulty of semantic masks covering the boundary regions of humans, which causes this information to contaminate the point cloud. The results for the low-dynamic sequence fr3/s/static are shown in \iref{fig:map-static}. The NGD-SLAM results show that the head information of the person on the left is completely leaked into the point cloud map, which is caused by the failure of object detection, while most of the information of the person on the right is completely retained in the point cloud map. Dyna-SLAM and DS-SLAM have removed most of the information of the two people, but from the side view, obvious contour noise blocks of people can be found, which is caused by the masks failing to fully cover the dynamic regions.
 
Overall, compared with the other three SLAM systems, GeneA-SLAM2 achieves the best effect in the global point cloud maps in the two sequences.

%%%%%%%%%%%%%%%%%%%%%%%%%%%%%%%%%%%%%%%%%%%%%%%%%%%%%%%%%%%%%%%%%%%%%%%%%%%%%%%%
\section{CONCLUSION}
\label{sec:conclusion}

This paper presents GeneA-SLAM2, an RGB-D SLAM system designed for dynamic environments, which aims to eliminate dynamic object interference through depth statistical information and improve the uniformity of keypoint distribution via a keypoints resampling algorithm. But due to the lenient mask mechanism, extremely small objects such as chair legs are missing. Experimental evaluations in a wide range of dynamic environments show that compared with baselines, GeneA-SLAM2 can obtain more accurate camera poses and cleaner global point cloud maps in dynamic environments. Our method mainly operates in indoor environments, and some parameters need to be obtained through statistics. In the future, automatic parameter calculation will be added. And the current construction is only point cloud maps with the dynamic regions removed. To enrich the environment representation, more advanced map modalities, such as semantic maps and object-level maps, should also be explored.

%%%%%%%%%%%%%%%%%%%%%%%%%%%%%%%%%%%%%%%%%%%%%%%%%%%%%%%%%%%%%%%%%%%%%%%%%%%%%%%%
% \section*{Limitation}

% Our method mainly operates in indoor environments, and some parameters need to be obtained through statistics. In the future, automatic parameter calculation will be added.

\section*{Acknowledgement}
This work is supported by the fund of Natural Science Fundamental Research Program of Shaanxi Province 2023-JC-QN-0645, 2023-JC-QN-0684 and 2024JC-YBQN-0645 and Xi'an Science and Technology Planning Project No.24NYGG0040.

% \addtolength{\textheight}{-10cm}   % This command serves to balance the column lengths

%%%%%%%%%%%%%%%%%%%%%%%%%%%%%%%%%%%%%%%%%%%%%%%%%%%%%%%%%%%%%%%%%%%%%%%%%%%%%%%%

%%%%%%%%%%%%%%%%%%%%%%%%%%%%%%%%%%%%%%%%%%%%%%%%%%%%%%%%%%%%%%%%%%%%%%%%%%%%%%%%

%%%%%%%%%%%%%%%%%%%%%%%%%%%%%%%%%%%%%%%%%%%%%%%%%%%%%%%%%%%%%%%%%%%%%%%%%%%%%%%%

%%%%%%%%%%%%%%%%%%%%%%%%%%%%%%%%%%%%%%%%%%%%%%%%%%%%%%%%%%%%%%%%%%%%%%%%%%%%%%%%

% References are important to the reader; therefore, each citation must be complete and correct. If at all possible, references should be commonly available publications.

\bibliographystyle{IEEEtran}
\bibliography{bibliography}

\begin{thebibliography}{10}
\providecommand{\url}[1]{#1}
\csname url@samestyle\endcsname
\providecommand{\newblock}{\relax}
\providecommand{\bibinfo}[2]{#2}
\providecommand{\BIBentrySTDinterwordspacing}{\spaceskip=0pt\relax}
\providecommand{\BIBentryALTinterwordstretchfactor}{4}
\providecommand{\BIBentryALTinterwordspacing}{\spaceskip=\fontdimen2\font plus
\BIBentryALTinterwordstretchfactor\fontdimen3\font minus \fontdimen4\font\relax}
\providecommand{\BIBforeignlanguage}[2]{{%
\expandafter\ifx\csname l@#1\endcsname\relax
\typeout{** WARNING: IEEEtran.bst: No hyphenation pattern has been}%
\typeout{** loaded for the language `#1'. Using the pattern for}%
\typeout{** the default language instead.}%
\else
\language=\csname l@#1\endcsname
\fi
#2}}
\providecommand{\BIBdecl}{\relax}
\BIBdecl

\bibitem{ORB3}
\BIBentryALTinterwordspacing
C.~Campos, R.~Elvira, J.~J.~G. Rodriguez, J.~M. M.~Montiel, and J.~D.~Tardos, ``Orb-slam3: An accurate open-source library for visual, visual–inertial, and multimap slam,'' \emph{IEEE Transactions on Robotics}, vol.~37, no.~6, p. 1874–1890, Dec. 2021. [Online]. Available: \url{http://dx.doi.org/10.1109/tro.2021.3075644}
\BIBentrySTDinterwordspacing

\bibitem{dyna-slam}
\BIBentryALTinterwordspacing
B.~Bescos, J.~M. Facil, J.~Civera, and J.~Neira, ``Dynaslam: Tracking, mapping, and inpainting in dynamic scenes,'' \emph{IEEE Robotics and Automation Letters}, vol.~3, no.~4, p. 4076–4083, Oct. 2018. [Online]. Available: \url{http://dx.doi.org/10.1109/lra.2018.2860039}
\BIBentrySTDinterwordspacing

\bibitem{ds-slam}
\BIBentryALTinterwordspacing
C.~Yu, Z.~Liu, X.-J. Liu, F.~Xie, Y.~Yang, Q.~Wei, and Q.~Fei, ``Ds-slam: A semantic visual slam towards dynamic environments,'' in \emph{2018 IEEE/RSJ International Conference on Intelligent Robots and Systems (IROS)}.\hskip 1em plus 0.5em minus 0.4em\relax IEEE, Oct. 2018, p. 1168–1174. [Online]. Available: \url{http://dx.doi.org/10.1109/iros.2018.8593691}
\BIBentrySTDinterwordspacing

\bibitem{cfp-slam}
\BIBentryALTinterwordspacing
X.~Hu, Y.~Zhang, Z.~Cao, R.~Ma, Y.~Wu, Z.~Deng, and W.~Sun, ``Cfp-slam: A real-time visual slam based on coarse-to-fine probability in dynamic environments,'' in \emph{2022 IEEE/RSJ International Conference on Intelligent Robots and Systems (IROS)}.\hskip 1em plus 0.5em minus 0.4em\relax IEEE, Oct. 2022, p. 4399–4406. [Online]. Available: \url{http://dx.doi.org/10.1109/iros47612.2022.9981826}
\BIBentrySTDinterwordspacing

\bibitem{ngdslam}
Y.~Zhang, M.~Bujanca, and M.~Luján, ``Ngd-slam: Towards real-time dynamic slam without gpu,'' \emph{arXiv preprint arXiv:2405.07392}, 2024.

\bibitem{d2slam}
\BIBentryALTinterwordspacing
A.~Beghdadi, M.~Mallem, and L.~Beji, ``D2slam: Semantic visual slam based on the depth-related influence on object interactions for dynamic environments,'' 2023. [Online]. Available: \url{https://arxiv.org/abs/2210.08647}
\BIBentrySTDinterwordspacing

\bibitem{dgslam}
\BIBentryALTinterwordspacing
Y.~Xu, H.~Jiang, Z.~Xiao, J.~Feng, and L.~Zhang, ``{DG}-{SLAM}: Robust dynamic gaussian splatting {SLAM} with hybrid pose optimization,'' in \emph{The Thirty-eighth Annual Conference on Neural Information Processing Systems}, 2024. [Online]. Available: \url{https://openreview.net/forum?id=tGozvLTDY3}
\BIBentrySTDinterwordspacing

\bibitem{genea-slam}
\BIBentryALTinterwordspacing
S.~Qing, A.~Li, J.~Liu, Y.~Gao, M.~Feng, F.~Nan, G.~Hu, J.~Wu, and Y.~Fan, ``Genea-slam: Enhancing slam with genetic algorithm-based feature points re-sampling,'' in \emph{2024 4th International Conference on Artificial Intelligence, Robotics, and Communication (ICAIRC)}.\hskip 1em plus 0.5em minus 0.4em\relax IEEE, Dec. 2024, p. 1042–1047. [Online]. Available: \url{http://dx.doi.org/10.1109/icairc64177.2024.10900093}
\BIBentrySTDinterwordspacing

\bibitem{ple-slam}
\BIBentryALTinterwordspacing
J.~He, M.~Li, Y.~Wang, and H.~Wang, ``Ple-slam: A visual-inertial slam based on point-line features and efficient imu initialization,'' \emph{IEEE Sensors Journal}, vol.~25, no.~4, p. 6801–6811, Feb. 2025. [Online]. Available: \url{http://dx.doi.org/10.1109/jsen.2024.3523039}
\BIBentrySTDinterwordspacing

\bibitem{silk-slam}
\BIBentryALTinterwordspacing
J.~Yao and Y.~Li, ``Silk-slam: accurate, robust and versatile visual slam with simple learned keypoints,'' \emph{Industrial Robot: the international journal of robotics research and application}, vol.~51, no.~3, p. 400–412, Mar. 2024. [Online]. Available: \url{http://dx.doi.org/10.1108/ir-11-2023-0309}
\BIBentrySTDinterwordspacing

\bibitem{blitz-slam}
\BIBentryALTinterwordspacing
Y.~Fan, Q.~Zhang, Y.~Tang, S.~Liu, and H.~Han, ``Blitz-slam: A semantic slam in dynamic environments,'' \emph{Pattern Recognition}, vol. 121, p. 108225, Jan. 2022. [Online]. Available: \url{http://dx.doi.org/10.1016/j.patcog.2021.108225}
\BIBentrySTDinterwordspacing

\bibitem{rds-slam}
\BIBentryALTinterwordspacing
Y.~Liu and J.~Miura, ``Rds-slam: Real-time dynamic slam using semantic segmentation methods,'' \emph{IEEE Access}, vol.~9, p. 23772–23785, 2021. [Online]. Available: \url{http://dx.doi.org/10.1109/access.2021.3050617}
\BIBentrySTDinterwordspacing

\bibitem{rodyn-slam}
\BIBentryALTinterwordspacing
H.~Jiang, Y.~Xu, K.~Li, J.~Feng, and L.~Zhang, ``Rodyn-slam: Robust dynamic dense rgb-d slam with neural radiance fields,'' \emph{IEEE Robotics and Automation Letters}, vol.~9, no.~9, p. 7509–7516, Sep. 2024. [Online]. Available: \url{http://dx.doi.org/10.1109/lra.2024.3427554}
\BIBentrySTDinterwordspacing

\bibitem{refusion}
\BIBentryALTinterwordspacing
E.~Palazzolo, J.~Behley, P.~Lottes, P.~Giguere, and C.~Stachniss, ``Refusion: 3d reconstruction in dynamic environments for rgb-d cameras exploiting residuals,'' in \emph{2019 IEEE/RSJ International Conference on Intelligent Robots and Systems (IROS)}.\hskip 1em plus 0.5em minus 0.4em\relax IEEE, Nov. 2019, p. 7855–7862. [Online]. Available: \url{http://dx.doi.org/10.1109/iros40897.2019.8967590}
\BIBentrySTDinterwordspacing

\bibitem{DBSCAN}
M.~Ester, H.-P. Kriegel, J.~Sander, and X.~Xu, ``A density-based algorithm for discovering clusters in large spatial databases with noise,'' in \emph{Proceedings of the Second International Conference on Knowledge Discovery and Data Mining}, ser. KDD'96.\hskip 1em plus 0.5em minus 0.4em\relax AAAI Press, 1996, p. 226–231.

\bibitem{autoencoder}
\BIBentryALTinterwordspacing
F.~Tian, B.~Gao, Q.~Cui, E.~Chen, and T.-Y. Liu, ``Learning deep representations for graph clustering,'' \emph{Proceedings of the AAAI Conference on Artificial Intelligence}, vol.~28, no.~1, Jun. 2014. [Online]. Available: \url{http://dx.doi.org/10.1609/aaai.v28i1.8916}
\BIBentrySTDinterwordspacing

\bibitem{opencv_library}
G.~Bradski, ``{The OpenCV Library},'' \emph{Dr. Dobb's Journal of Software Tools}, 2000.

\bibitem{tum}
\BIBentryALTinterwordspacing
J.~Sturm, N.~Engelhard, F.~Endres, W.~Burgard, and D.~Cremers, ``A benchmark for the evaluation of rgb-d slam systems,'' in \emph{2012 IEEE/RSJ International Conference on Intelligent Robots and Systems}.\hskip 1em plus 0.5em minus 0.4em\relax IEEE, Oct. 2012, p. 573–580. [Online]. Available: \url{http://dx.doi.org/10.1109/iros.2012.6385773}
\BIBentrySTDinterwordspacing

\bibitem{tele-slam}
\BIBentryALTinterwordspacing
T.~Ji, C.~Wang, and L.~Xie, ``Towards real-time semantic rgb-d slam in dynamic environments,'' in \emph{2021 IEEE International Conference on Robotics and Automation (ICRA)}.\hskip 1em plus 0.5em minus 0.4em\relax IEEE, May 2021, p. 11175–11181. [Online]. Available: \url{http://dx.doi.org/10.1109/icra48506.2021.9561743}
\BIBentrySTDinterwordspacing

\bibitem{acefusion}
\BIBentryALTinterwordspacing
M.~Bujanca, B.~Lennox, and M.~Luján, ``Acefusion - accelerated and energy-efficient semantic 3d reconstruction of dynamic scenes,'' in \emph{2022 IEEE/RSJ International Conference on Intelligent Robots and Systems (IROS)}.\hskip 1em plus 0.5em minus 0.4em\relax IEEE, Oct. 2022, p. 11063–11070. [Online]. Available: \url{http://dx.doi.org/10.1109/iros47612.2022.9981591}
\BIBentrySTDinterwordspacing

\bibitem{evo}
M.~Grupp, ``evo: Python package for the evaluation of odometry and slam,'' \url{https://github.com/MichaelGrupp/evo}, 2017.

\bibitem{umeyama}
\BIBentryALTinterwordspacing
S.~Umeyama, ``Least-squares estimation of transformation parameters between two point patterns,'' \emph{IEEE Transactions on Pattern Analysis and Machine Intelligence}, vol.~13, no.~4, p. 376–380, Apr. 1991. [Online]. Available: \url{http://dx.doi.org/10.1109/34.88573}
\BIBentrySTDinterwordspacing

\end{thebibliography}

\end{document}